\documentclass{article}

\usepackage{microtype}
\usepackage{graphicx}
\usepackage{subcaption}
\usepackage{booktabs} 

\usepackage{hyperref}

\usepackage[preprint]{icml2026}

\usepackage{amsmath}
\usepackage{amssymb}
\usepackage{mathtools}
\usepackage{amsthm}

\usepackage[capitalize,noabbrev]{cleveref}

\theoremstyle{plain}

\theoremstyle{definition}

\theoremstyle{remark}

\usepackage[textsize=tiny]{todonotes}

\begin{document}

\twocolumn[
  \icmltitle{Transformers Learn Robust In-Context Regression \\ under Distributional Uncertainty}



  \icmlsetsymbol{equal}{*}

  \begin{icmlauthorlist}
    \icmlauthor{Hoang T.H. Cao}{hcmut}
    \icmlauthor{Hai D.V. Trinh}{hcmut}
    \icmlauthor{Tho Quan}{hcmut}
    \icmlauthor{Lan V. Truong}{hcmut}
  \end{icmlauthorlist}

  \icmlaffiliation{hcmut}{
    Faculty of Computer Science and Engineering, \\
    Ho Chi Minh University of Technology (HCMUT), \\
    Vietnam National University Ho Chi Minh City (VNU-HCM), Vietnam
    }

  \icmlcorrespondingauthor{Lan V. Truong}{lantv@hcmut.edu.vn}

  \icmlkeywords{Machine Learning, ICML}

  \vskip 0.3in
]



\printAffiliationsAndNotice{}  

\begin{abstract}
  Recent work has shown that Transformers can perform in-context learning for linear regression under restrictive assumptions, including i.i.d. data, Gaussian noise, and Gaussian regression coefficients. However, real-world data often violate these assumptions: the distributions of inputs, noise, and coefficients are typically unknown, non-Gaussian, and may exhibit dependency across the prompt. This raises a fundamental question: can Transformers learn effectively in-context under realistic distributional uncertainty? We study in-context learning for noisy linear regression under a broad range of distributional shifts, including non-Gaussian coefficients, heavy-tailed noise, and non-i.i.d. prompts. We compare Transformers against classical baselines that are optimal or suboptimal under the corresponding maximum-likelihood criteria. Across all settings, Transformers consistently match or outperform these baselines, demonstrating robust in-context adaptation beyond classical estimators.
\end{abstract}

\section{Introduction}

\subsection{Motivation}

Transformers have increasingly demonstrated the ability to perform
\emph{in-context learning} (ICL): given a short prompt consisting of
input--output pairs, the model infers an underlying task and predicts
the output for a new query without updating its parameters.
Understanding what algorithm a Transformer implicitly executes in
context, and under which conditions this algorithm is effective, is
important for both theoretical analysis and practical deployment.
Prior work~\cite{garg2022what} shows that for linear regression tasks
with i.i.d.\ Gaussian features and Gaussian noise, Transformer-based
ICL often aligns closely with ordinary least squares (OLS), achieving
competitive performance using only the examples provided in the
prompt. This observation motivates a compelling interpretation of
Transformers as amortizing classical estimators through sequence
modeling.

Recently,~\cite{shi2024exploring} demonstrated that Transformers can retain strong in-context learning (ICL) performance under various forms of label noise $\varepsilon$. They report that under different noise distributions (Gaussian, Uniform, Exponential, and Poisson), the Transformer outperforms other baseline models. However, their evaluation relies solely on MSE as the performance metric, which introduces bias in two respects: (1) some baselines are not optimal when the noise is non-Gaussian, and (2) the Transformer may be inherently favored due to its training with a least-squares loss. Moreover, their analysis is restricted to settings in which both the features $\mathbf{x}$ and the regression coefficients $\mathbf{w}$ follow canonical Gaussian priors, typically under noiseless or highly simplified regression models.

In contrast, real-world data rarely satisfy such assumptions. Features may exhibit temporal dependencies, cross-dimensional correlations, non-stationary drift, or structured latent geometry, while the underlying coefficients are often sparse, heavy-tailed, or drawn from non-Gaussian priors. In these regimes, classical OLS is known to be brittle, and alternative estimators—such as Ridge regression, Lasso, or Bayesian methods tailored to the data geometry—can achieve substantially improved performance.

If Transformers implicitly behave like OLS when trained on i.i.d.\
Gaussian tasks, it is therefore unclear whether this inductive bias
persists, adapts, or becomes detrimental when the structure of $\mathbf{x}$ or
$\mathbf{w}$ deviates from the Gaussian i.i.d.\ regime. In other words, does
in-context learning inherit the sensitivity of OLS to feature geometry,
or can Transformers exhibit emergent robustness and adaptation to
nontrivial feature and coefficient distributions?

This paper addresses this question through a systematic study of
Transformer-based in-context learning for linear regression under
distributional shifts in both the features and the regression
coefficients. We vary the distributions of $\mathbf{x}$, $\mathbf{w}$ and $\varepsilon$ away from the
standard i.i.d.\ Gaussian setting and compare Transformer performance
against classical baselines, including OLS and Ridge regression, under
matched priors. We further evaluate both in-distribution performance
and generalization under shifts in feature or coefficient structure.
By isolating changes in $\mathbf{x}$, $\mathbf{w}$ and $\varepsilon$, our experiments reveal whether
Transformers merely implement Bayes estimation in context or
exhibit robustness and adaptation beyond suboptimal baselines, and identify the regimes
---in terms of sample size, dimensionality, feature covariance, and
coefficient structure---under which such behaviors emerge.

\subsection{Related Work}

\paragraph{In-Context Learning.}
The phenomenon of in-context learning was first highlighted by
\cite{brown2020language_models_few_shot}, who demonstrated that GPT-3
can perform new tasks from a small number of examples provided in the
prompt, without any parameter updates. Since then, a growing body of
work has sought to understand the mechanisms and limitations of this
capability \cite{chen2022transinr,agarwal2024manyshot,lu2024emergent,ren2024representation,anwar2025understanding}. Several studies have shown that Transformers can perform
ICL on simple or low-dimensional function classes, including linear
models~\cite{DaiSunDongHaoMaSuiWei2022,garg2022what,BaiChenWangXiongMei2023,zhang2024trained}.
Other lines of work interpret ICL as a form of implicit Bayesian
inference~\cite{XieRaghunathanLiMa2021,Reuter2025}, or as implementing
gradient-based optimization within the forward pass of the
Transformer~\cite{Ahn2023,vonOswaldNiklassonRandazzoSacramentoMordvintsevZhmoginovVladymyrov2022}.
More recently, \cite{WangJiangLi2024} show that when Transformers are
explicitly constructed to approximate gradient descent, they can
support in-context learning across a broad range of function classes
beyond linear regression. At the same time,
\cite{Hill2025} argue that effective ICL critically depends
on the match between the pre-training and test distributions, and that
Transformers do not in general implement a universal least-squares
procedure. In contrast to these works, our focus is not on identifying
the precise internal mechanism of ICL, but rather on characterizing
its performance under varying distributions of features $\mathbf{x}$,
coefficients $\mathbf{w}$, and noise $\varepsilon$.
\paragraph{Beyond Gaussian Distributions.}
While prior studies~\cite{XieRaghunathanLiMa2021,WeiTayTranWebsonLuChenLiuHuangZhouMa2023}
demonstrate strong ICL performance for linear regression tasks,
including inverse regression settings~\cite{LuYu2025}, these analyses
largely assume Gaussian distributions for features, coefficients, and
noise. Recent work suggests that deviations from Gaussian noise can
lead to qualitatively different ICL behavior~\cite{shi2024exploring},
and that even mild distributional shifts can substantially degrade
Transformer performance~\cite{Hill2025}. Our work contributes to this
line of research by systematically examining Transformer-based ICL
under non-Gaussian, non i.i.d. feature and coefficient distributions, and by
comparing its behavior against classical estimators that are optimal
or suboptimal under the corresponding assumptions.

\subsection{Contributions}

This work makes the following contributions:

\paragraph{Systematic study of in-context learning under distributional uncertainty.}
  We conduct a comprehensive empirical investigation of Transformer-based
  in-context learning for linear regression when the distributions of
  features $\mathbf{x}$, regression coefficients $\mathbf{w}$, and noise
  $\varepsilon$ deviate from the standard i.i.d.\ Gaussian assumptions.
  Our study isolates the effect of each source of distributional shift,
  enabling a fine-grained characterization of when and how in-context
  learning succeeds.

  \paragraph{Evaluation against ML-optimal and suboptimal classical baselines.}
  For each noise model considered, we compare Transformers against
  classical estimators that are optimal or suboptimal under the
  corresponding maximum-likelihood criterion, including least squares,
  Ridge regression, and $\ell_1$-based estimators. This allows us to assess
  whether Transformers merely reproduce classical solutions in context or
  exhibit behavior beyond fixed estimators.

  \paragraph{Evidence of robust in-context adaptation beyond ordinary least squares.}
  Across a broad range of non-Gaussian coefficient priors, structured and
  non-i.i.d.\ feature distributions, and heavy-tailed or discrete noise
  models, we find that Transformers consistently match optimal or outperform
  suboptimal classical baselines.
  These results suggest that in-context learning enables robust adaptation
  to distributional misspecification rather than simply implementing
  ordinary least squares in context.

\section{Model Settings}
\noindent We study the in-context learning capabilities of Transformer models on various function classes derived from the Noisy Linear Regression Model. Our setup closely follows the methodology established by \cite{garg2022what} but is significantly extended to incorporate non-Gaussian distributions for noise and coefficients, and non-i.i.d. features.

\subsection{Linear Regression Task}
We study in-context learning on a standard linear model with additive noise. For an input matrix \(X \in \mathbb{R}^{k \times d}\) (rows are examples) and an unknown coefficient vector \(w \in \mathbb{R}^d\)
, labels are generated as
\begin{equation}
y = Xw + \varepsilon
\label{eq:linear_model}
\end{equation}
where \(\varepsilon\) is additive noise drawn from either a Gaussian distribution or a specified non-Gaussian family. Unless otherwise stated, features are i.i.d. with zero mean and unit variance, and \(w\) is sampled once per task instance from a chosen prior (Gaussian by default). The prompt length (number of in-context examples) is 
\(k\) , the ambient dimension is \(d\), and the query is the 
\( (k+1) \)-st example.

Each evaluation presents the Transformer with a single sequence that interleaves inputs and outputs from the same task instance:
\[
\underbrace{(x_1,y_1),\ (x_2,y_2),\ \ldots,\ (x_k,y_k)}_{\text{prefix of }k\text{ labeled examples}},\ \ x_{k+1}
\]
The model receives the entire prefix of labeled examples together with the \emph{unlabeled} query
input \(x_{k+1}\) and must produce \(\hat y_{k+1}\). Concretely, both inputs \(x_i \in \mathbb{R}^d\) and outputs \(y_i \in \mathbb{R}\) are mapped into the model's embedding space using a single shared linear projection. 
For each labeled example, the scalar \(y_i\) is expanded into a \(d\)-dimensional vector by placing \(y_i\) in the first coordinate and zeros elsewhere, allowing \(x_i\) and \(y_i\) to share the same read-in layer. 
The resulting representations are interleaved to form the in-context sequence
\[
[x_1,\ y_1,\ x_2,\ y_2,\ \ldots,\ x_k,\ y_k,\ x_{k+1}].
\]
Predictions are read out at the query positions (i.e., immediately after each \(x_{t+1}\)),
and the training/evaluation loss is the mean-squared error (MSE) at those query positions.
\subsection{Coefficients ($\mathbf{w}$), Features ($\mathbf{X}$), and Noise ($\varepsilon$)}
To rigorously evaluate the adaptive capacity of the Transformer-based estimator, we systematically vary the statistical regimes of the components in Equation~\ref{eq:linear_model}. Unless stated otherwise, noise scales are calibrated to achieve a fixed target variance—whenever the variance is finite—in order to ensure comparability in terms of Signal-to-Noise Ratio (SNR) across experimental settings.

\subsubsection{Coefficient Distributions ($\mathbf{w}$)}

The coefficient prior plays a central role in determining both identifiability and estimator bias in linear regression. To probe the Transformer's sensitivity to different structural priors—and to evaluate whether it adapts its in-context behavior accordingly—we consider the following families of coefficient distributions.
\paragraph{Laplace distribution.}  
    Each coefficient is sampled independently from a Laplace distribution. This setting preserves symmetry around zero while inducing heavier tails than the Gaussian prior, serving as a minimal deviation from the standard assumption. It allows us to test whether the Transformer retains optimal regression behavior when Gaussianity is relaxed but rotational symmetry is approximately preserved.

    \paragraph{Exponential distribution.}  
    Coefficients are drawn independently from an Exponential distribution, $\mathbf{w}_i \sim \mathrm{Exp}(\lambda)$, resulting in non-negative and highly skewed magnitudes. This prior introduces strong structural constraints—namely positivity and magnitude imbalance—which are not naturally exploited by classical OLS or Ridge estimators. This regime is particularly informative in the underdetermined setting ($k < d$), where prior information can significantly affect estimation quality.
    \paragraph{Uniform hypersphere distribution.}
Coefficients are sampled uniformly from the unit hypersphere in
$\mathbb{R}^d$, i.e., $\mathbf{w} \sim \mathrm{Unif}(\mathbb{S}^{d-1})$.
This prior enforces a fixed $\ell_2$ norm while preserving full rotational
symmetry and isotropy.
Unlike the Laplace or exponential priors, it does not induce sparsity,
skewness, or coordinate-wise structure, and therefore serves as a
geometric control case for assessing sensitivity to coefficient structure.

\subsubsection{Feature (Covariate) Distributions ($\mathbf{x}$)}

To evaluate the effect of feature distribution priors on in-context learning,
we vary the distribution of input covariates $\mathbf{x}$ while keeping the
regression coefficients and noise fixed.
Unlike the standard i.i.d.\ Gaussian setting, we consider feature distributions
that induce nontrivial marginal structure or dependency across context steps.
All models are trained and evaluated in-distribution.

Specifically, we consider the following feature distributions.

\paragraph{Gamma Features.}
Each feature dimension is sampled independently from a Gamma distribution.
For each context step $t$ and dimension $j$, we draw
\begin{equation}
\mathbf{x}_{t,j} \sim \mathrm{Gamma}(\alpha, \theta),
\end{equation}
where $\alpha > 0$ is the concentration (shape) parameter and $\theta > 0$
is the rate parameter.
This distribution produces strictly non-negative, skewed features with
unbounded support, departing significantly from symmetric Gaussian inputs.
Feature dimensions and context steps are sampled independently under this
setting.

\paragraph{VAR(1) Features.}
To introduce temporal dependence across the context, we consider features
generated by a vector autoregressive process of order one.
The feature sequence $\{\mathbf{x}_t\}_{t=1}^{k}$ is generated as
\begin{align}
\mathbf{x}_1 &\sim \mathcal{N}(\mathbf{0}, I), \\
\mathbf{x}_t &= A \mathbf{x}_{t-1} + \boldsymbol{\varepsilon}_t,
\quad t = 2, \dots, k,
\end{align}
where $A \in \mathbb{R}^{d \times d}$ is the transition matrix and
$\boldsymbol{\varepsilon}_t \sim \mathcal{N}(\mathbf{0}, \sigma^2 I)$
is i.i.d.\ Gaussian innovation noise.
In our experiments, $A$ is chosen to be a scaled identity matrix,
$A = \rho I$, with $\rho \in (0,1)$ controlling the strength of temporal
correlation.
This construction induces strong dependence across context steps while
preserving marginal Gaussianity at each time step.

Unless otherwise specified, no truncation or dimensional masking is applied
to the feature vectors, and all feature dimensions are observed by the model.

\subsubsection{Noise Distributions ($\varepsilon$)}

We consider five noise distributions for the observation noise
$\varepsilon$ in the linear model
\begin{equation}
y_i = \mathbf{w}^\top \mathbf{x}_i + \varepsilon_i,
\quad i \in [n].
\end{equation}
For each noise model, we characterize the corresponding
maximum-likelihood (ML) optimal loss whenever it admits a tractable form.
When the variance of the noise distribution is finite, parameters are
chosen such that the resulting noise variance matches a fixed target
value to ensure comparable signal-to-noise ratios across settings.

\paragraph{Bernoulli Noise.}
We consider Bernoulli noise with parameter $p \in (0, 1/2]$,
\begin{equation}
\varepsilon_i \sim \mathrm{Bernoulli}(p).
\end{equation}
As derived in Appendix~\ref{sec:ml_losses}, the ML-optimal estimator under
this noise model corresponds to minimizing the $\ell_1$ loss,
\begin{equation}
\hat{\mathbf{w}}
=
\arg\min_{\mathbf{w} \in \mathbb{R}^d}
\sum_{i=1}^{n}
\left| y_i - \mathbf{w}^\top \mathbf{x}_i \right|.
\end{equation}
Accordingly, we include $\ell_1$-based baselines, solved via linear
programming and ADMM, when evaluating performance under Bernoulli noise.

\paragraph{Exponential Noise.}
We consider one-sided exponential noise with rate parameter $\lambda > 0$,
\begin{equation}
\varepsilon_i \sim \mathrm{Exponential}(\lambda),
\end{equation}
with support on $[0, \infty)$.
As shown in Appendix~\ref{sec:ml_losses}, the corresponding ML-optimal
objective again reduces to $\ell_1$ regression,
\begin{equation}
\hat{\mathbf{w}}
=
\arg\min_{\mathbf{w} \in \mathbb{R}^d}
\sum_{i=1}^{n}
\left| y_i - \mathbf{w}^\top \mathbf{x}_i \right|.
\end{equation}
This setting allows us to test whether Transformers implicitly adapt to
a non-quadratic loss when the noise distribution is highly skewed.

\paragraph{Gamma Noise.}
We further consider Gamma-distributed noise with shape $\alpha > 0$ and
scale $\theta > 0$,
\begin{equation}
\varepsilon_i \sim \mathrm{Gamma}(\alpha, \theta).
\end{equation}
The corresponding log-likelihood yields the ML objective
\begin{equation}
\hat{\mathbf{w}}
=
\arg\max_{\mathbf{w} \in \mathbb{R}^d}
(\alpha - 1)
\sum_{i=1}^{n}
\log \left| y_i - \mathbf{w}^\top \mathbf{x}_i \right|
\end{equation}
\begin{equation}
-
\frac{1}{\theta}
\sum_{i=1}^{n}
\left| y_i - \mathbf{w}^\top \mathbf{x}_i \right|,
\end{equation}
which is non-convex and does not admit a closed-form solution.
In this case, we primarily compare against $\ell_1$-based baselines as a
tractable surrogate, following common practice.

\paragraph{Poisson Noise.}
For discrete noise, we consider Poisson-distributed observations with
rate $\lambda > 0$,
\begin{equation}
\varepsilon_i \sim \mathrm{Poisson}(\lambda).
\end{equation}
The resulting likelihood leads to an ML objective involving factorial
terms,
\begin{equation}
\prod_{i=1}^{n}
\frac{\lambda^{\, y_i - \mathbf{w}^\top \mathbf{x}_i}}
{(y_i - \mathbf{w}^\top \mathbf{x}_i)!},
\end{equation}
which does not yield a tractable optimization problem for $\mathbf{w}$.
We therefore retain squared-error-based baselines in this setting and
use it as a stress test for robustness under discrete, non-Gaussian
noise.

\paragraph{Student-$t$ Noise.}
Finally, we consider heavy-tailed Student-$t$ noise with two degrees of
freedom,
\begin{equation}
\varepsilon_i \sim t_{\nu}.
\end{equation}
This distribution has infinite variance ($\nu \leq 2$) and induces extreme outliers.
While the likelihood is well defined, the corresponding ML objective
does not admit a closed-form loss and is highly non-convex.
We therefore do not include an explicit ML-optimal baseline for this
case, and instead use it to evaluate the robustness of Transformer-based
in-context learning under severe model misspecification.

\section{Experimental Results}
\subsection{Experimental Setup}
In this section, we present empirical results evaluating Transformer-based
in-context learning under distributional shifts in the regression
coefficients, feature distributions, and observation noise.
Unless otherwise stated, all results are reported for models trained
for 500{,}000 optimization steps using the setup described in Appendix~\ref{sec:ex_setup}.
\subsection{Coefficient Distribution Shift}

We investigate the effect of non-Gaussian and structured coefficient priors on
in-context learning.
Throughout this section, the feature distribution is held fixed.
Unless otherwise specified, observation noise is either absent or drawn
from a zero-mean Gaussian distribution with small variance, allowing us
to isolate the influence of the coefficient prior.
All models are trained using an $\ell_2$ objective and evaluated under the same
loss.

\begin{figure}[t]
  \centering
  \begin{subfigure}{0.48\columnwidth}
    \includegraphics[width=\linewidth]{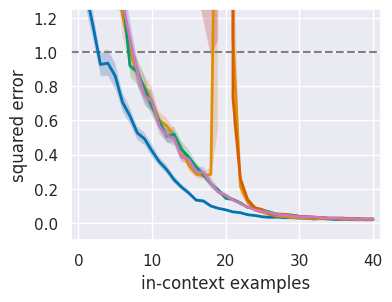}
    \caption{Exponential prior ($\lambda=1.0$)}
  \end{subfigure}
  \hfill
  \begin{subfigure}{0.48\columnwidth}
    \includegraphics[width=\linewidth]{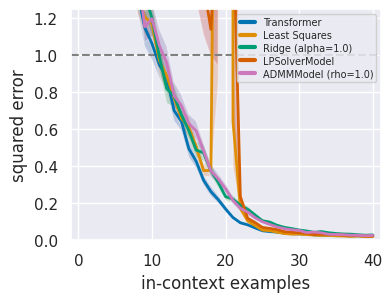}
    \caption{Laplace prior ($b = 1.0$)}
  \end{subfigure}

  \vspace{0.8em}

  \begin{subfigure}{0.8\columnwidth}
    \centering
    \includegraphics[width=\linewidth]{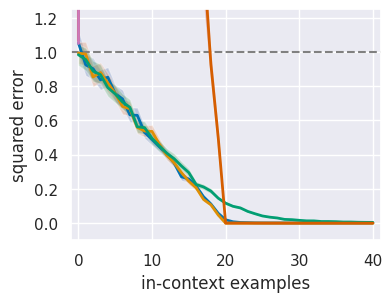}
    \caption{Uniform hypersphere prior ($\|w\|_2 = 1$)}
  \end{subfigure}

  \caption{
    In-context prediction error under different coefficient priors in the
    noiseless setting.
    All methods are trained with $\ell_2$ loss.
    Top row: exponential (left) and Laplace (right) coefficient distributions.
    Bottom row: coefficients sampled uniformly from the unit hypersphere.
    We compare Transformer-based in-context learning against OLS, Ridge, and
    $\ell_1$-based solvers (LP and ADMM).
    Errors are reported as normalized excess loss relative to a
    curriculum-dependent baseline.
  }
  \label{fig:w_shift}
\end{figure}

\paragraph{Discussion.}
Figure~\ref{fig:w_shift} demonstrates that Transformer-based in-context learning
is highly sensitive to the structure of the coefficient prior, even when all
models are trained and evaluated under the same $\ell_2$ objective.

For exponentially distributed coefficients (Fig.~\ref{fig:w_shift}, top left),
the Transformer consistently outperforms OLS, Ridge, and $\ell_1$-based solvers
for small to moderate context lengths ($k \leq 30$).
The exponential prior is asymmetric, supported on the non-negative orthant, and
induces strong anisotropy in parameter space.
Classical $\ell_2$ estimators implicitly assume symmetry around zero and are
therefore statistically mismatched in this regime, while $\ell_1$ estimators are
not optimal in the absence of Laplace-type structure.
The substantial finite-sample advantage of the Transformer indicates that its
in-context behavior adapts to the structural bias imposed by the coefficient
distribution, rather than implementing a fixed estimator.

For Laplace-distributed coefficients (Fig.~\ref{fig:w_shift}, top right), the
Transformer again achieves lower error in the low-to-moderate context regime, but
the performance gap narrows as the number of in-context examples increases.
When $k \gtrsim 25$, all methods converge to comparable performance.
This behavior is consistent with the symmetry and finite variance of the Laplace
prior, under which $\ell_2$-based estimators become asymptotically efficient.
In this setting, the Transformer's advantage appears to be primarily a
finite-sample effect.

In contrast, when coefficients are sampled uniformly from the unit hypersphere
(Fig.~\ref{fig:w_shift}, bottom), we consider a noiseless setting.
In this case, the Transformer closely tracks OLS, Ridge, and
$\ell_1$ baselines across all context lengths.
This prior preserves rotational symmetry and isotropy, despite imposing a global
norm constraint.
As a result, no estimator can exploit directional bias or sparsity in the
coefficient distribution.
The absence of a systematic performance gap in this case serves as a sanity
check: when the coefficient prior aligns with the implicit assumptions of
classical estimators, Transformer-based in-context learning does not exhibit
spurious gains.

Beyond predictive performance, these results provide evidence that the
Transformer implicitly infers distributional properties of the coefficient
prior from the in-context examples.
In particular, the systematic advantages observed under skewed or asymmetric
priors suggest that the model is not merely fitting labels conditioned on the
features, but is forming an internal representation of salient characteristics
of the coefficient distribution, such as asymmetry, support constraints, or
effective sparsity.

From this perspective, Transformer-based in-context learning can be interpreted
as performing a form of implicit prior inference.
Given a short context, the model appears to identify the underlying coefficient
distribution and to adapt its effective inductive bias accordingly.
This behavior is qualitatively different from that of classical estimators, whose
inductive biases are fixed \emph{a priori} and do not adapt to the realized prior
at test time.

Importantly, this implicit inference emerges despite the Transformer being
trained with a fixed $\ell_2$ objective and without explicit access to the
coefficient distribution.
This suggests that the model does not implement a single estimator, but rather
approximates a family of estimators indexed by the coefficient prior.
In this sense, the Transformer acts as a distribution-aware meta-estimator,
interpolating between classical solutions depending on the structure of the
in-context data.

Finally, since all errors are reported as normalized excess loss relative to a
curriculum-based baseline that accounts for the effective rank of the design
matrix, the improvements observed under exponential priors reflect genuine gains
in statistical efficiency, rather than artifacts of scaling or normalization.


\subsection{Feature Distribution Shift}

Next, we examine the effect of feature distribution shifts on in-context
learning, focusing on deviations from the standard i.i.d.\ Gaussian
assumption.
In contrast to coefficient or noise misspecification, changes in the
feature distribution directly affect the geometry of the design matrix
$\mathbf{x}$, and hence the conditioning and identifiability properties of the
underlying regression problem.
We consider two representative settings: i.i.d.\ non-Gaussian features
and temporally correlated features across the prompt.
In all experiments in this section, additive observation noise is present
and follows a zero-mean Gaussian distribution with fixed variance.

\begin{figure}[htbp]
  \centering
  \begin{subfigure}{0.48\columnwidth}
    \includegraphics[width=\linewidth]{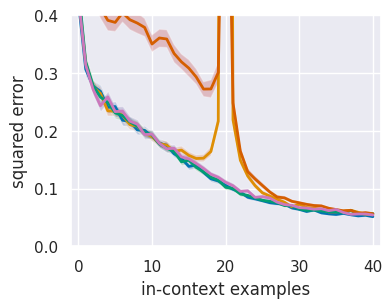}
    \caption{Gamma-distributed features ($\alpha = 2.0$, $\theta = 1.0$)}
  \end{subfigure}
  \hfill
  \begin{subfigure}{0.48\columnwidth}
    \includegraphics[width=\linewidth]{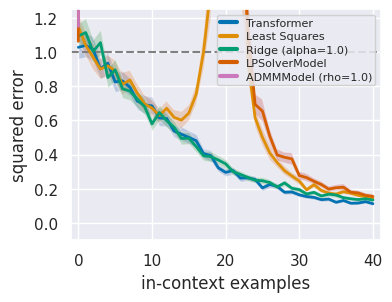}
    \caption{Correlated features (VAR(1)) $\rho=0.4$}

  \end{subfigure}
  \caption{
    Effect of feature distribution shifts on in-context learning.
    Gamma features induce skewed, non-negative marginals while preserving
    independence across context steps.
    VAR(1) features introduce temporal dependence across the prompt through
    an autoregressive structure.
  }
  \label{fig:x_shift}
\end{figure}

\paragraph{Discussion.}
Figure~\ref{fig:x_shift} illustrates that the impact of feature distribution
shifts depends critically on whether such shifts alter the fundamental
identifiability of the linear regression problem.
When features are sampled i.i.d.\ from a Gamma distribution, all methods—
including least squares, Ridge regression, and the Transformer—exhibit
nearly identical performance.
This behavior is expected: under independence, additive noise, and finite
second moments, ordinary least squares remains statistically consistent
regardless of the specific marginal distribution of the covariates.
Accordingly, no method enjoys a systematic advantage in this regime.

In contrast, temporally correlated features generated by a VAR(1) process
introduce structured dependence across the prompt.
Although the resulting design matrix remains full-rank, correlations across
context steps degrade the effective conditioning of the regression problem
in the presence of additive noise, and violate the implicit i.i.d.\
assumptions underlying classical estimators.
In this setting, we observe that the Transformer closely tracks the
performance of OLS and related baselines, neither collapsing nor exhibiting
instability.
This indicates that in-context learning remains well-behaved under moderate
temporal dependence, but does not exploit VAR(1) structure in a way that
yields a decisive advantage over linear estimators.

Taken together, these results suggest that feature distribution shifts alone
do not induce the same robustness gaps observed under coefficient or noise
misspecification.
Instead, the benefits of Transformer-based in-context learning appear most
pronounced when the statistical structure within the prompt alters the
effective loss landscape or introduces information that is not readily
captured by second-order feature statistics alone.

\subsection{Noise Distribution Shift}

Finally, we evaluate robustness to different observation noise models.
These experiments compare Transformer performance against baselines that
are optimal or suboptimal under the corresponding maximum-likelihood
criteria.
\begin{figure}[t]
  \centering

  \begin{subfigure}{0.48\columnwidth}
    \centering
    \includegraphics[width=\linewidth]{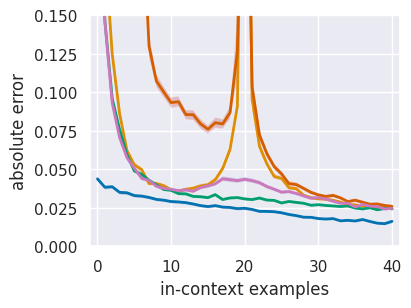}
    \caption{Bernoulli noise ($p = 0.25$)}
  \end{subfigure}
  \hfill
  \begin{subfigure}{0.48\columnwidth}
    \centering
    \includegraphics[width=\linewidth]{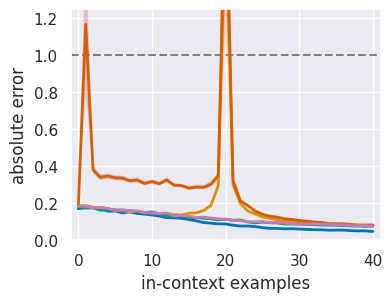}
    \caption{Exponential noise $\lambda=1.0$)}
  \end{subfigure}

  \vspace{0.6em}

  \begin{subfigure}{0.48\columnwidth}
    \centering
    \includegraphics[width=\linewidth]{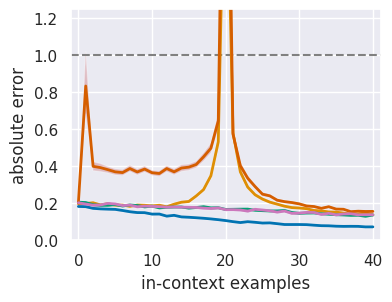}
    \caption{Gamma noise ($\alpha = 2.0$, $\theta = 1.0$)}
  \end{subfigure}
  \hfill
  \begin{subfigure}{0.48\columnwidth}
    \centering
    \includegraphics[width=\linewidth]{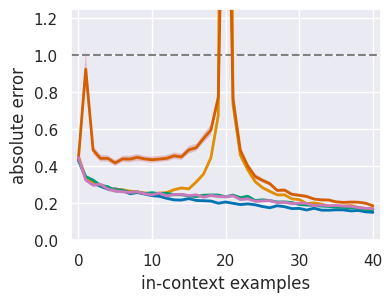}
    \caption{Poisson noise ($\lambda = 1.0$)}
  \end{subfigure}

  \vspace{0.6em}

  \begin{subfigure}{0.9\columnwidth}
    \centering
    \includegraphics[width=1\linewidth]{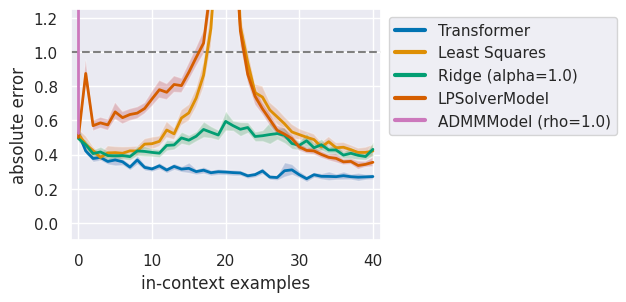}
    \caption{Student-$t$ noise ($\nu = 2$)}
  \end{subfigure}

  \caption{
Robustness of in-context learning under non-Gaussian and heavy-tailed
noise distributions.
All models in this figure are trained and evaluated using $\ell_1$ loss.
We compare Transformer-based in-context learning with classical estimators,
including least squares, Ridge regression, and $\ell_1$ solvers (LP and ADMM).
For Bernoulli, exponential, Gamma, and Poisson noise, the $\ell_1$ objective
corresponds to the maximum-likelihood estimator.
The Student-$t$ case ($\nu = 2$) is shown separately, as it admits no
finite-variance likelihood and falls outside the classical ML framework.
}

  \label{fig:e_shift}
\end{figure}

\paragraph{Discussion.}
Figure~\ref{fig:e_shift} evaluates the robustness of in-context learning
under a range of non-Gaussian noise models, all trained and evaluated
using an $\ell_1$ loss.
Across all settings, we observe that Transformer-based in-context learning
matches the performance of classical estimators, including
those that are maximum-likelihood optimal under the assumed noise model.

\textbf{Bernoulli noise.}
In the Bernoulli case, all baselines achieve very low error, reflecting
the bounded and discrete nature of the noise.
Nevertheless, the Transformer consistently attains slightly lower error
across all context lengths.
While classical $\ell_1$ estimators are statistically optimal in this
setting, they tend to transition sharply between underdetermined
($k < d$) and overdetermined ($k > d$) regimes.
The Transformer's smoother performance suggests an implicit regularization
effect that stabilizes estimation across the interpolation threshold.

\textbf{Exponential noise.}
For exponential noise, where the ML-optimal estimator corresponds to
$\ell_1$ regression with one-sided support, the Transformer does not
exhibit a strong advantage in the severely underdetermined regime.
However, once the context length slightly exceeds the dimensionality
($k \gtrsim d$), it consistently outperforms all baselines.
This behavior is consistent with the asymmetry of the noise distribution:
classical estimators require sufficient samples to resolve the skewed
likelihood, whereas the Transformer appears to exploit distributional
structure more efficiently in the moderate-sample regime.

\textbf{Gamma noise.}
Under Gamma noise, the Transformer shows a clear and persistent advantage
across the full range of context lengths.
Although the Gamma likelihood induces a non-convex log-likelihood due to
the logarithmic term, the $\ell_1$ objective serves as a principled and
robust surrogate.
The Transformer's superior performance in this regime suggests that it
implements an effective estimator that goes beyond fixed-form convex
approximations, adapting to the heavy-tailed, positive-support structure
of the noise.

\textbf{Poisson noise.}
In the Poisson setting, where the exact ML estimator lacks a closed-form
solution, all baselines exhibit relatively flat error curves.
The Transformer achieves modest but consistent improvements once
$k \gtrsim 10$, indicating that it can partially compensate for the
misspecification inherent in both $\ell_2$ and $\ell_1$ baselines.
This suggests that in-context learning may approximate aspects of the
Poisson likelihood that are not captured by standard linear estimators.

\textbf{Student-$t$ noise.}
The Student-$t$ case ($\nu = 2$) presents the most challenging regime,
as the noise distribution has infinite variance and admits no
maximum-likelihood estimator under finite-moment assumptions.
In this setting, least squares and Ridge regression fail to improve with
additional context, while only the Transformer and, to a lesser extent,
ADMM-based $\ell_1$ regression exhibit decreasing error.
The Transformer's clear advantage highlights its robustness to extreme
heavy-tailed noise and indicates that its in-context behavior cannot be
explained by any classical estimator tied to a fixed loss or likelihood.

Overall, these results demonstrate that Transformer-based in-context
learning does not merely recover ML-optimal estimators when they exist,
but continues to perform effectively even in regimes where classical
statistical theory provides no optimal solution.
This robustness underscores the ability of Transformers to adapt their
effective in-context estimation strategy to the noise structure, rather
than committing to a single predefined loss.

\section{Conclusion}

Our results provide a unified empirical perspective on in-context learning
for linear regression under distributional uncertainty. By systematically
varying the distributions of the regression coefficients $\mathbf{w}$,
features $\mathbf{x}$, and noise $\varepsilon$, we show that Transformer-based
in-context learning exhibits a level of robustness that is difficult to
explain as merely implementing ordinary least squares (OLS) within the
forward pass.

\paragraph{Beyond Ordinary Least Squares.}
A central takeaway of our study is that, although prior work has shown
that Transformers can match OLS performance under i.i.d.\ Gaussian
assumptions, this behavior does not fully characterize in-context learning.
Across non-Gaussian coefficient priors, structured and non-i.i.d.\ feature
distributions, and heavy-tailed or discrete noise models, Transformers
consistently match or outperform OLS and Ridge regression.
These findings suggest that in-context learning is not rigidly tied to a
single estimator, but instead adapts to the statistical structure present
within the prompt.

\paragraph{Implicit Adaptation to Noise and Loss Structure.}
Our experiments with non-Gaussian noise highlight a particularly striking
phenomenon: despite being trained using squared-error objectives,
Transformers often perform comparably to estimators optimized for alternative
loss functions, such as $\ell_1$ loss under Bernoulli or exponential noise.
This behavior is especially pronounced in heavy-tailed regimes, including
Gamma and Student-$t$ noise, where classical estimators either require
carefully matched loss functions or fail to improve with additional data.
These results suggest that in-context learning implicitly adapts its
effective loss to the noise distribution observed in the prompt, rather
than adhering strictly to the training objective.
While our experiments do not identify the underlying mechanism, they
demonstrate that the effective estimator implemented by the Transformer
is data-dependent and distribution-aware.

\paragraph{Sensitivity to Feature Geometry.}
We further find that Transformers are substantially less sensitive to
feature geometry than classical linear estimators.
In settings with correlated, anisotropic, or non-Gaussian features—where
the conditioning of the design matrix degrades OLS performance—
Transformers maintain strong in-context generalization.
This suggests that in-context learning may exploit sequential structure
to encode feature geometry implicitly, rather than relying on explicit
matrix inversion or fixed regularization schemes.

\paragraph{Robustness Versus Computational Cost.}
While the empirical robustness of Transformer-based in-context learning
is compelling, it comes with an important computational trade-off.
The attention mechanism underlying Transformers scales quadratically
with the prompt length, incurring an $O(n^2)$ cost in the number of
in-context examples.
In contrast, classical estimators such as OLS, Ridge regression, or
$\ell_1$ solvers admit closed-form solutions or iterative algorithms with
significantly lower computational complexity.
From a practical standpoint, this suggests that Transformer-based
in-context learning is most attractive in regimes where robustness to
distributional misspecification is critical and the cost of incorrect
model assumptions outweighs the increased computational burden.
In settings where the data-generating process is well understood and
classical assumptions hold, simpler estimators may remain preferable.

\paragraph{Implications for In-Context Learning Theory.}
Taken together, our results complement recent theoretical and mechanistic
accounts of in-context learning.
Rather than contradicting interpretations of ICL as implicit optimization
or Bayesian inference, our findings emphasize that the effective behavior
of in-context learning depends critically on the distributional structure
present at test time.
Any comprehensive theory of in-context learning must therefore account
not only for how Transformers implement learning in context, but also for
why this learning remains robust under substantial distributional
misspecification.

\paragraph{Limitations and Future Directions.}
Our study is limited to linear regression tasks, and it remains an open
question whether similar robustness properties extend to nonlinear
function classes or more complex prediction settings.
Moreover, while we compare against classical estimators that are optimal
under known distributions, we do not attempt to characterize the precise
algorithm executed by the Transformer in context.
Future work may seek to bridge this gap by connecting distributional
robustness with mechanistic analyses of attention dynamics, or by
investigating whether similar robustness can be achieved with models
that admit more favorable computational scaling.

Overall, our results suggest that Transformer-based in-context learning
provides a flexible and robust alternative to classical estimators when
the underlying data-generating process is unknown or misspecified.
At the same time, they highlight an inherent trade-off between robustness
and computational efficiency that should be carefully considered in
practical deployments.

\nocite{langley00}

\bibliography{example_paper}
\bibliographystyle{icml2026}

\newpage
\appendix
\onecolumn


\section{Experimental Setup} \label{sec:ex_setup}

\subsection{Model Architecture and Training Setup}

We adopt architectures from the GPT-2 family throughout all experiments.
Unless otherwise specified, all models share the same configuration,
consisting of an embedding dimension of 256, 12 Transformer layers, and
8 attention heads. Models are trained for a total of 500{,}000 optimization
steps using a batch size of 64, with the learning rate fixed at
$3 \times 10^{-4}$.

The model $\mathcal{M}_\theta$ is trained to minimize the expected loss over
a distribution of prompts $\mathcal{P}$ according to
\begin{equation}
\min_{\theta}
\mathbb{E}_{\mathbf{P} \sim \mathcal{P}}
\left[
\ell\big(
\mathcal{M}_\theta(\mathbf{P}_{k, \mathbf{x}_{k+1}}),
y_{k+1}
\big)
\right],
\end{equation}
where $\ell(\cdot,\cdot)$ denotes the squared error loss and
$\mathbf{P}_{k, \mathbf{x}_{k+1}}$ denotes the input sequence
$[\mathbf{x}_1, y_1, \ldots, \mathbf{x}_k, y_k, \mathbf{x}_{k+1}]$.
This training procedure casts the Transformer as a meta-learner that
optimizes its ability to adapt in-context across a distribution of linear
regression tasks.

\paragraph{Curriculum Learning.}Natural function classes typically contain functions of varying complexity. We exploit this structure by training the model using a curriculum ~\cite{bengio2009curriculum,kong2021adaptive,soviany2022curriculum}, starting from a simplified function distribution (e.g., linear functions whose weight vectors lie in a low-dimensional subspace) and progressively increasing the complexity. This curriculum significantly speeds up training and, in many cases, makes it feasible to train models that would be considerably more costly without such a strategy

To accelerate training and improve optimization stability, we adopt a
curriculum learning strategy in which the model is first exposed to
simpler prompt configurations and gradually transitioned to more complex
ones. During early training stages, prompt inputs $\mathbf{x}_i$ are
restricted to lie in a lower-dimensional subspace, and each prompt
contains fewer input--output pairs. As training progresses, both the
subspace dimensionality and the prompt length are increased according to
a fixed schedule.

Concretely, at each curriculum stage, we zero out all but the first
$d_{\mathrm{cur}}$ coordinates of $\mathbf{x}_i$ and sample prompts
consisting of $k_{\mathrm{cur}}$ input--output pairs. Training begins with
$d_{\mathrm{cur}} = 5$ and $k_{\mathrm{cur}} = 11$, and every 2{,}000
optimization steps, $d_{\mathrm{cur}}$ and $k_{\mathrm{cur}}$ are increased
by 1 and 2, respectively, until reaching the full setting
$d_{\mathrm{cur}} = d$ and $k_{\mathrm{cur}} = 2d + 1$.

\paragraph{Training Resources.}
All models are trained on a single NVIDIA GeForce RTX 4070 Super GPU.
With curriculum learning enabled, each training run requires
approximately 12--16 hours. When training is conducted using a fixed
low-dimensional setting ($d = 5$, $k = 11$), the total training time is
reduced to approximately 4--5 hours. For rapid experimentation, we
further reduce the number of training steps to 50{,}000.

\subsection{Baselines}

\paragraph{Least Squares.}
Given a prompt
\[
\mathbf{P} = (\mathbf{x}_1, y_1, \ldots, \mathbf{x}_k, y_k, \mathbf{x}_{\mathrm{query}}),
\]
we construct a design matrix $X \in \mathbb{R}^{k \times d}$ whose $i$-th row
is $\mathbf{x}_i^\top$, and a response vector $y \in \mathbb{R}^k$.
The minimum-norm least-squares estimator is
\begin{equation}
\hat{\mathbf{w}} = X^{+} y,
\end{equation}
where $X^{+}$ denotes the Moore--Penrose pseudoinverse. The prediction is
given by
\begin{equation}
M(\mathbf{P}) = \hat{\mathbf{w}}^\top \mathbf{x}_{\mathrm{query}}.
\end{equation}

\paragraph{Ridge Regression.}
Ridge regression introduces $\ell_2$ regularization to stabilize estimation.
The estimator is
\begin{equation}
\hat{\mathbf{w}}_{\mathrm{ridge}} =
\left(X^\top X + \alpha I\right)^{-1} X^\top y,
\end{equation}
where $\alpha > 0$ is the regularization coefficient. The prediction is
\begin{equation}
M(\mathbf{P}) = \hat{\mathbf{w}}_{\mathrm{ridge}}^\top
\mathbf{x}_{\mathrm{query}}.
\end{equation}


\paragraph{$\ell_1$ Regression (Linear Programming Solver).}
For several noise distributions considered in this work—most notably
Bernoulli, exponential, and Laplace noise—the maximum-likelihood estimator
corresponds to minimizing the $\ell_1$ loss.
We therefore include $\ell_1$ regression as a baseline whenever such a
maximum-likelihood interpretation is valid.

Concretely, given a prompt with design matrix $X \in \mathbb{R}^{k \times d}$
and responses $y \in \mathbb{R}^k$, the estimator is defined as
\begin{equation}
\hat{\mathbf{w}}_{\ell_1}
=
\arg\min_{\mathbf{w} \in \mathbb{R}^d}
\sum_{i=1}^k
\lvert y_i - \mathbf{w}^\top \mathbf{x}_i \rvert.
\end{equation}
This objective is convex and can be formulated as a linear program by
introducing auxiliary variables that represent the positive and negative
parts of the residuals.
In our experiments, we solve this problem using a standard linear
programming solver.The resulting estimator is treated as the ML-optimal baseline for the
corresponding noise models.



\paragraph{$\ell_1$ Regression via ADMM.}
In addition to a generic linear programming solver, we implement an
$\ell_1$ regression baseline using the alternating direction method of
multipliers (ADMM).
This approach provides an alternative optimization strategy for the same
$\ell_1$ objective and is commonly used in large-scale or structured
settings where generic LP solvers may be inefficient or numerically
unstable.

Introducing an auxiliary variable $\mathbf{z}$ to decouple the residuals,
the problem can be written as
\begin{align}
\min_{\mathbf{w}, \mathbf{z}} \quad
& \sum_{i=1}^k |z_i| \\
\text{s.t.} \quad
& \mathbf{z} = y - X \mathbf{w}.
\end{align}
ADMM alternates between closed-form updates of $\mathbf{w}$, soft-thresholding
updates of $\mathbf{z}$, and dual variable updates until convergence or a
maximum number of iterations is reached.
We include this baseline to verify that observed performance trends are
not artifacts of a particular solver, but are consistent across different
implementations of $\ell_1$ regression.

For both $\ell_1$ baselines, predictions for the query input are obtained
as
\begin{equation}
M(\mathbf{P}) = \hat{\mathbf{w}}^\top \mathbf{x}_{\mathrm{query}}.
\end{equation}

\section{Maximum-Likelihood Optimal Estimators} \label{sec:ml_losses}

We consider the linear model
\begin{equation}
y_i = \mathbf{w}^\top \mathbf{x}_i + \varepsilon_i,
\quad \forall i \in [n].
\end{equation}
Assume that $\varepsilon_1, \varepsilon_2, \dots, \varepsilon_n$ are i.i.d.,
and that $P_{\varepsilon}$ denotes the common distribution of the noise
variables. Then,
\begin{equation}
P(y_i \mid \mathbf{x}_i, \mathbf{w})
=
P_{\varepsilon}\!\left(y_i - \mathbf{w}^\top \mathbf{x}_i\right),
\quad \forall i \in [n].
\end{equation}
It follows that the likelihood is
\begin{align}
P(\mathbf{y} \mid \mathbf{x}, \mathbf{w})
&=
\prod_{i=1}^{n} P(y_i \mid \mathbf{x}_i, \mathbf{w}) \\
&=
\prod_{i=1}^{n}
P_{\varepsilon}\!\left(y_i - \mathbf{w}^\top \mathbf{x}_i\right).
\end{align}
The maximum-likelihood estimator of $\mathbf{w}$ is therefore
\begin{equation}
\hat{\mathbf{w}}
= \arg\max_{\mathbf{w} \in \mathbb{R}^n} P(\mathbf{y}\mid \mathbf{x}, \mathbf{w})
= \arg\max_{\mathbf{w} \in \mathbb{R}^n}
\prod_{i=1}^{n} P_{\varepsilon}\!\left(y_i - \mathbf{w}^\top \mathbf{x}_i\right).
\end{equation}
\subsection{Bernoulli Noise}

Assume that $\varepsilon_1, \varepsilon_2, \dots, \varepsilon_n$ are i.i.d.\
Bernoulli random variables with parameter $p \in [0, 1/2]$.
Then,
\begin{equation} 
P_{\varepsilon}(y)
=
p^{y}(1-p)^{1-y}
=
\left(\frac{p}{1-p}\right)^{y}(1-p),
\quad \forall y \in \{0,1\}.
\end{equation}
The ML-optimal estimator is
\begin{align}
\hat{\mathbf{w}}
&=
\arg\max_{\mathbf{w} \in \mathbb{R}^d}
\prod_{i=1}^{n}
P_{\varepsilon}\!\left(y_i - \mathbf{w}^\top \mathbf{x}_i\right) \\
&=
\arg\max_{\mathbf{w} \in \mathbb{R}^d}
\prod_{i=1}^{n}
\left(\frac{p}{1-p}\right)^{y_i - \mathbf{w}^\top \mathbf{x}_i}(1-p) \\
&=
\arg\max_{\mathbf{w} \in \mathbb{R}^d}
\sum_{i=1}^{n}
\left(y_i - \mathbf{w}^\top \mathbf{x}_i\right)
\log\!\left(\frac{p}{1-p}\right) \label{eq:loglikelihood} \\
&=
-\arg\max_{\mathbf{w} \in \mathbb{R}^d}
\sum_{i=1}^{n}
\left(y_i - \mathbf{w}^\top \mathbf{x}_i\right) \label{eq:34} \\
&=
\arg\min_{\mathbf{w} \in \mathbb{R}^d}
\sum_{i=1}^{n}
\left|y_i - \mathbf{w}^\top \mathbf{x}_i\right|. \label{eq:35}
\end{align}
where \eqref{eq:34} follows from $p \in [0, 1/2]$, and \eqref{eq:35} follows from the fact that
$y_i - \mathbf{w}^\top \mathbf{x}_i = \varepsilon_i \in \{0,1\}$.

\subsection{Exponential Noise}

Assume that $\varepsilon_1, \varepsilon_2, \dots, \varepsilon_n$ are i.i.d.\
exponential random variables with parameter $\lambda > 0$.
Then,
\begin{equation}
P_{\varepsilon}(y;\lambda)
=
\lambda e^{-\lambda y}\mathbf{1}\{y \ge 0\},
\quad \forall y \in \mathbb{R}.
\end{equation}
The ML-optimal estimator is
\begin{align}
\hat{\mathbf{w}}
&=
\arg\max_{\mathbf{w} \in \mathbb{R}^d}
\prod_{i=1}^{n}
P_{\varepsilon}\!\left(y_i - \mathbf{w}^\top \mathbf{x}_i\right) \\
&=
\arg\max_{\mathbf{w} \in \mathbb{R}^d}
\prod_{i=1}^{n}
\lambda e^{-\lambda (y_i - \mathbf{w}^\top \mathbf{x}_i)}
\mathbf{1}\{y_i - \mathbf{w}^\top \mathbf{x}_i \ge 0\} \\
&=
\arg\max_{\mathbf{w} \in \mathbb{R}^d}
\prod_{i=1}^{n}
\lambda e^{-\lambda \left|y_i - \mathbf{w}^\top \mathbf{x}_i\right|}
\mathbf{1}\{y_i - \mathbf{w}^\top \mathbf{x}_i \ge 0\} \\
&= \label{eq:40}
\arg\min_{\mathbf{w} \in \mathbb{R}^d}
\sum_{i=1}^{n}
\left|y_i - \mathbf{w}^\top \mathbf{x}_i\right|.
\end{align}
where \eqref{eq:40} follows from the fact that $y_i - \mathbf{w}^\top \mathbf{x}_i = \varepsilon_i \geq 0 $ for all $i \in [n]$.
\subsection{Gamma Noise}

Assume that $\varepsilon_1, \varepsilon_2, \dots, \varepsilon_n$ are i.i.d.\
Gamma random variables with parameters $\alpha > 0$ and $\theta > 0$.
Then,
\begin{equation}
P_{\varepsilon}(y;\alpha,\theta)
=
\frac{1}{\Gamma(\alpha)\theta^{\alpha}}
y^{\alpha-1}e^{-y/\theta}
\mathbf{1}\{y \ge 0\}.
\end{equation}
The ML-optimal estimator is
\begin{align}
\hat{\mathbf{w}}
&=
\arg\max_{\mathbf{w} \in \mathbb{R}^d}
\prod_{i=1}^{n}
P_{\varepsilon}\!\left(y_i - \mathbf{w}^\top \mathbf{x}_i\right)
\tag{18}
\\[0.5em]
&=
\arg\max_{\mathbf{w} \in \mathbb{R}^d}
\prod_{i=1}^{n}
\frac{1}{\Gamma(\alpha)\,\theta^{\alpha}}
\left(y_i - \mathbf{w}^\top \mathbf{x}_i\right)^{\alpha-1}
\exp\!\left(-\frac{y_i - \mathbf{w}^\top \mathbf{x}_i}{\theta}\right)
\mathbf{1}\!\left\{y_i - \mathbf{w}^\top \mathbf{x}_i \ge 0\right\}
\tag{19}
\\[0.5em]
&=
\arg\max_{\mathbf{w} \in \mathbb{R}^d}
(\alpha - 1)
\sum_{i=1}^{n}
\log\!\left|y_i - \mathbf{w}^\top \mathbf{x}_i\right|
-
\frac{1}{\theta}
\sum_{i=1}^{n}
\left|y_i - \mathbf{w}^\top \mathbf{x}_i\right|.
\tag{20}
\end{align}

\subsection{Poisson Noise}

Assume that $\varepsilon_1, \varepsilon_2, \dots, \varepsilon_n$ are i.i.d.\
Poisson random variables with parameter $\lambda > 0$.
Then,
\begin{equation}
P_{\varepsilon}(y;\lambda)
=
e^{-\lambda}\frac{\lambda^{y}}{y!},
\quad \forall y \in \mathbb{N}.
\end{equation}
The ML-optimal estimator is
\begin{align}
\hat{\mathbf{w}}
&=
\arg\max_{\mathbf{w} \in \mathbb{R}^d}
\prod_{i=1}^{n}
P_{\varepsilon}\!\left(y_i - \mathbf{w}^\top \mathbf{x}_i\right) \\
&=
\arg\max_{\mathbf{w} \in \mathbb{R}^d}
\prod_{i=1}^{n}
e^{-\lambda}
\frac{\lambda^{y_i - \mathbf{w}^\top \mathbf{x}_i}}
{(y_i - \mathbf{w}^\top \mathbf{x}_i)!},
\end{align}
which does not admit a closed-form solution.

\section{Additional Experiments}
\label{app:additional_experiments}

This appendix extends the main experimental results by varying key
distributional parameters in Figures~1--3.
All experiments here reuse the same experimental setups as in the main
paper and differ only in the parameters of the coefficient, feature, or
noise distributions.
Unless otherwise stated, models are trained for $200{,}000$ optimization
steps.
The purpose of these experiments is to verify that the observed trends
are robust to reasonable parameter variations, rather than artifacts of
a specific configuration.
\subsection{Extension of Figure~2: Feature Distribution Parameters}
\label{app:fig2_extension}

We extend Figure~2 by varying the parameters of the feature distributions
considered in the main paper, while keeping the coefficient distribution
and noise model fixed. To isolate the effect of the feature distribution itself, we do not add observation noise to the output $y$ in this experiment, allowing a clearer comparison across the four feature settings.
We consider the following $(\text{concentration}, \text{rate})$ settings:
$(2,2)$, $(3,0.5)$, $(4,2)$, and $(5,1)$.

\begin{figure}[H]
  \centering
  \begin{subfigure}{0.24\columnwidth}
    \includegraphics[width=\linewidth]{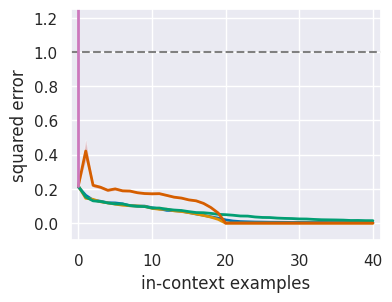}
    \caption{$(2,2)$}
  \end{subfigure}
  \hfill
  \begin{subfigure}{0.24\columnwidth}
    \includegraphics[width=\linewidth]{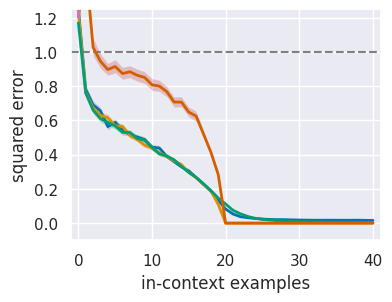}
    \caption{$(3,0.5)$}
  \end{subfigure}
  \hfill
  \begin{subfigure}{0.24\columnwidth}
    \includegraphics[width=\linewidth]{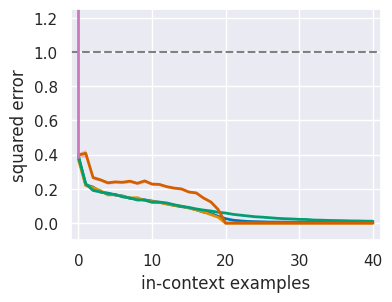}
    \caption{$(4,2)$}
  \end{subfigure}
  \hfill
  \begin{subfigure}{0.24\columnwidth}
    \includegraphics[width=\linewidth]{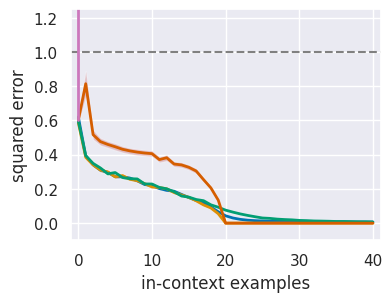}
    \caption{$(5,1)$}
  \end{subfigure}

  \caption{
    Extension of Figure~2 (Gamma-distributed features).
    In-context learning performance across different Gamma shape and rate
    parameters.
  }
  \label{fig:app_fig2_gamma}
\end{figure}
\paragraph{Gamma features}
Figure~\ref{fig:app_fig2_gamma} shows that varying the shape and rate parameters
of the Gamma feature distribution does not lead to qualitative differences in
performance.
Across all settings, Transformer-based in-context learning closely tracks OLS
and other classical baselines, achieving comparable prediction error.
Despite changes in feature marginal skewness and concentration, no systematic
advantage emerges, indicating that feature distribution shifts alone—when
independence is preserved—are insufficient to induce performance gaps between
in-context learning and linear estimators.

\paragraph{VAR(1) correlated features (diagonal AR(1) structure).}
We consider a special case of the VAR(1) model in which the transition
matrix is diagonal, $A = \rho I$, inducing identical AR(1)-type dynamics
along each feature dimension.
This setting isolates temporal correlation effects while avoiding
cross-feature interactions.
We examine the effect of feature correlation by varying the autoregressive
coefficient $\rho \in \{0.2, 0.5, 0.8\}$.
As $\rho$ increases, temporal dependence across the prompt becomes stronger,
altering the effective covariance structure of the design matrix
and reducing the number of effectively independent samples. To isolate the effect of feature correlation, no observation noise is added to the output $y$ in this experiment.

Across all correlation strengths, Transformer-based in-context learning remains
stable and achieves performance comparable to that of classical baselines.
Notably, this holds even in regimes where standard estimators are
expected to suffer from multicollinearity.
Although stronger correlations reduce the effective sample diversity, the
Transformer consistently adapts to the induced dependence structure and
maintains low prediction error.
Overall, varying $\rho$ does not lead to qualitative changes in performance,
indicating that the Transformer effectively accommodates correlated features
without degradation relative to standard estimators.

\begin{figure}[H]
  \centering
  \begin{subfigure}{0.32\columnwidth}
    \includegraphics[width=\linewidth]{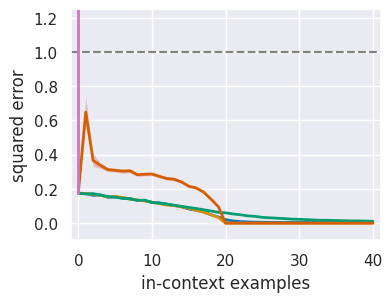}
    \caption{$\rho=0.2$}
  \end{subfigure}
  \hfill
  \begin{subfigure}{0.32\columnwidth}
    \includegraphics[width=\linewidth]{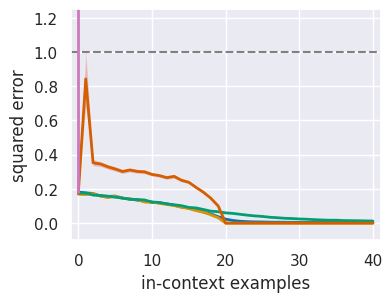}
    \caption{$\rho=0.5$}
  \end{subfigure}
  \hfill
  \begin{subfigure}{0.32\columnwidth}
    \includegraphics[width=\linewidth]{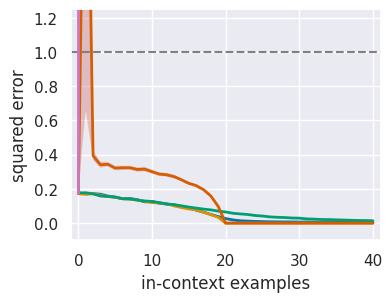}
    \caption{$\rho=0.8$}
  \end{subfigure}

  \caption{
    Extension of Figure~2 (VAR(1) features).
    Effect of increasing temporal correlation strength across the prompt.
  }
  \label{fig:app_fig2_var1}
\end{figure}

\subsection{Extension of Figure~3: Noise Distribution Parameters}
\label{app:fig3_extension}

\begin{figure}[H]
  \centering
  \begin{subfigure}{0.24\columnwidth}
    \includegraphics[width=\linewidth]{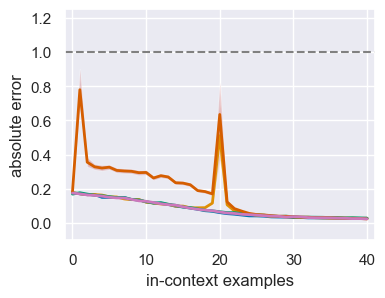}
    \caption{Bernoulli $p=0.1$}
  \end{subfigure}
  \hfill
  \begin{subfigure}{0.24\columnwidth}
    \includegraphics[width=\linewidth]{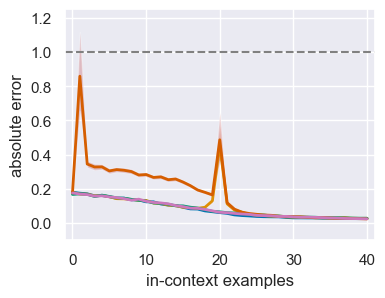}
    \caption{Bernoulli $p=0.2$}
  \end{subfigure}
  \hfill
  \begin{subfigure}{0.24\columnwidth}
    \includegraphics[width=\linewidth]{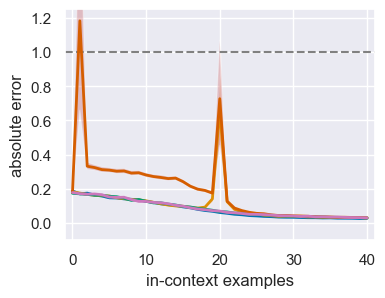}
    \caption{Bernoulli $p=0.3$}
  \end{subfigure}
  \hfill
  \begin{subfigure}{0.24\columnwidth}
    \includegraphics[width=\linewidth]{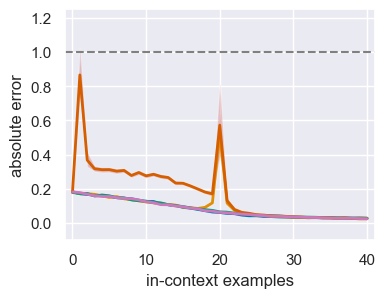}
    \caption{Bernoulli $p=0.4$}
  \end{subfigure}

  \caption{
    Extension of Figure~3 (noise parameter sweeps).
    Performance under varying Bernoulli, Gamma, Poisson, and Student-$t$
    noise parameters.
    The qualitative trends observed in the main paper persist across
    a broad range of noise intensities.
  }
  \label{fig:app_fig3_noise}
\end{figure}

We extend Figure~3 by varying the parameters of the noise distributions
considered in the main paper, while keeping the feature and coefficient
distributions fixed.
Varying the noise parameters induces corresponding changes in the effective
signal-to-noise ratio (SNR), allowing us to assess the robustness of
in-context learning across a range of noise intensities and distributional
regimes.
\paragraph{Bernoulli noise.}
Figure~\ref{fig:app_fig3_noise} reports performance under Bernoulli noise with
varying success probability $p \in \{0.1, 0.2, 0.3, 0.4\}$.
Across all values of $p$, Transformer-based in-context learning remains stable
and achieves error levels comparable to the strongest classical baselines.
In all cases, the loss decreases rapidly with increasing context length and
converges to a small value, indicating effective adaptation to discrete noise.

Notably, the qualitative behavior of the learning curves is highly consistent
across different values of $p$.
While increasing $p$ rescales the noise magnitude, it does not fundamentally
alter the structure of the estimation problem.
As a result, both the Transformer and classical estimators exhibit similar
convergence patterns, and no pronounced sensitivity to the Bernoulli parameter
is observed.
These results suggest that, in contrast to heavy-tailed continuous noise, the
Transformer's performance under Bernoulli perturbations is largely invariant to
moderate changes in noise intensity.

\paragraph{Exponential noise.}
We vary the rate parameter $\lambda \in \{0.5, 1.5, 2.0\}$, interpolating between
more heavy-tailed and less skewed exponential noise distributions.
As $\lambda$ increases, the effective noise variance decreases and the
estimation problem becomes progressively better conditioned.

At intermediate rates (e.g., $\lambda = 2.0$), $\ell_1$-based baselines such as
LPSolver and ADMM exhibit rapid convergence, leading to lower error when the
number of in-context examples is close to the ambient dimension ($k \approx d$).
This behavior is consistent with the improved stability of convex $\ell_1$
estimators under less heavy-tailed noise.

Across all values of $\lambda$, the Transformer maintains stable performance and
closely matches the strongest baseline methods.
Moreover, its error decreases smoothly as $\lambda$ increases, indicating
robustness to changes in tail behavior and effective noise scale.
Unlike classical estimators, whose performance is tightly coupled to specific
distributional assumptions, the Transformer adapts continuously across noise
regimes without requiring explicit tuning to the noise parameter.

\begin{figure}[H]
  \centering
  \begin{subfigure}{0.32\columnwidth}
    \includegraphics[width=\linewidth]{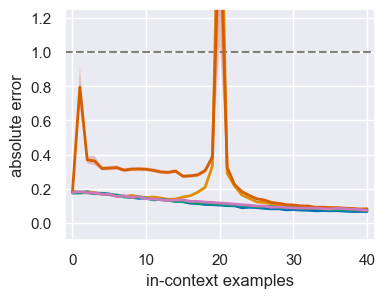}
    \caption{$\lambda=0.5$}
  \end{subfigure}
  \hfill
  \begin{subfigure}{0.32\columnwidth}
    \includegraphics[width=\linewidth]{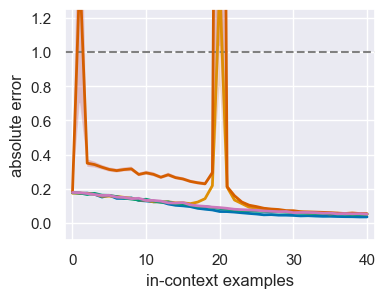}
    \caption{$\lambda=1.5$}
  \end{subfigure}
  \hfill
  \begin{subfigure}{0.32\columnwidth}
    \includegraphics[width=\linewidth]{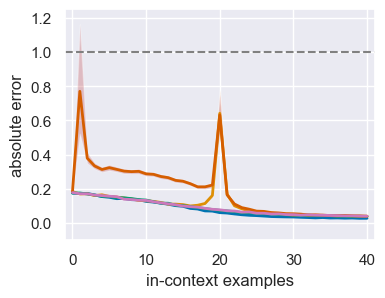}
    \caption{$\lambda=2.0$}
  \end{subfigure}

  \caption{
  Extension of Figure~3 (Exponential noise).
  Performance under exponential noise with varying rate parameter $\lambda$.
}

  \label{fig:app_fig3_exponential}
\end{figure}
\paragraph{Gamma noise.}
We evaluate Gamma-distributed noise under multiple $(k,\lambda)$ configurations.
For $(k=2,\lambda=2)$, the Transformer exhibits a mild but consistent advantage
over classical baselines once the number of in-context examples exceeds the
ambient dimension.
As the shape parameter increases to $(k=3,\lambda=1)$, the performance gap
becomes more pronounced, with the Transformer outperforming all baselines across
a broader range of context lengths.
This trend further strengthens for $(k=4,\lambda=1)$, where the Transformer
achieves the lowest error overall.
These results indicate that increasing concentration of the Gamma noise
amplifies the Transformer's ability to leverage additional in-context examples.

\begin{figure}[H]
  \centering
  \begin{subfigure}{0.32\columnwidth}
    \includegraphics[width=\linewidth]{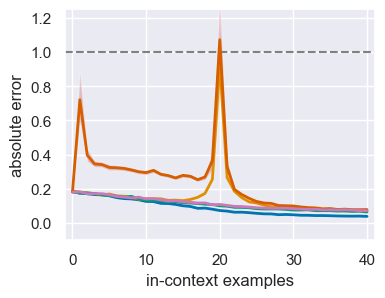}
    \caption{$(\alpha=2,\theta=2)$}
  \end{subfigure}
  \hfill
  \begin{subfigure}{0.32\columnwidth}
    \includegraphics[width=\linewidth]{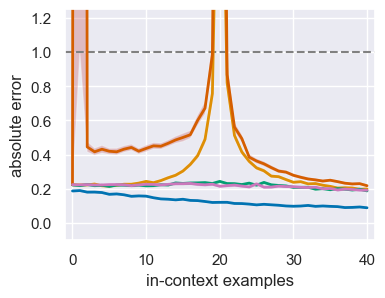}
    \caption{$(\alpha=3,\theta=1)$}
  \end{subfigure}
  \hfill
  \begin{subfigure}{0.32\columnwidth}
    \includegraphics[width=\linewidth]{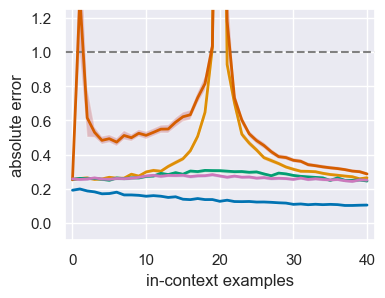}
    \caption{$(\alpha=4,\theta=1)$}
  \end{subfigure}

  \caption{
    Extension of Figure~3 (Gamma noise).
    In-context learning under Gamma-distributed noise with varying
    shape and rate parameters.
  }
  \label{fig:app_fig3_gamma}
\end{figure}

\paragraph{Poisson noise.}
We vary the Poisson rate parameter $\lambda \in \{0.5, 2.0, 3.0\}$, corresponding
to increasing noise variance and stronger discreteness.
For small rates ($\lambda = 0.5$), Transformer-based in-context learning closely
matches classical baselines across all context lengths.

As $\lambda$ increases, the difference between the Transformer and classical
methods becomes more pronounced.
Across all settings, classical baselines exhibit a characteristic saturation
behavior: when the number of in-context examples satisfies $k < d$, their
performance remains largely flat, reflecting insufficient effective rank in the
design matrix.
When $k > d$, the baselines converge to similar error levels and show little
additional improvement.

In contrast, the Transformer continues to reduce loss beyond the $k \approx d$
threshold, achieving progressively larger gains as $\lambda$ increases.
This behavior suggests that the Transformer leverages additional in-context
examples to adapt its effective estimator under increasing noise variance,
rather than relying on a fixed parametric solution.

\begin{figure}[H]
  \centering
  \begin{subfigure}{0.32\columnwidth}
    \includegraphics[width=\linewidth]{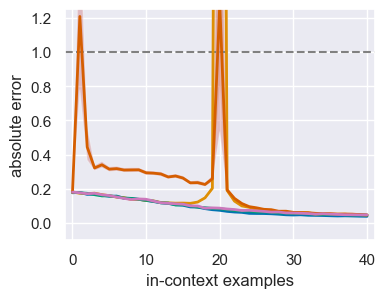}
    \caption{$\lambda=0.5$}
  \end{subfigure}
  \hfill
  \begin{subfigure}{0.32\columnwidth}
    \includegraphics[width=\linewidth]{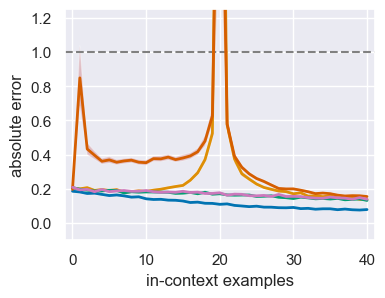}
    \caption{$\lambda=2.0$}
  \end{subfigure}
  \hfill
  \begin{subfigure}{0.32\columnwidth}
    \includegraphics[width=\linewidth]{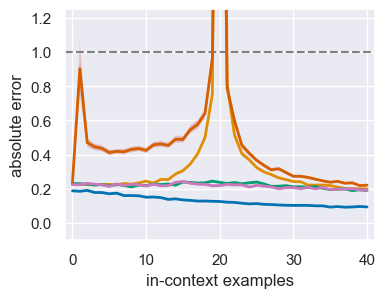}
    \caption{$\lambda=3.0$}
  \end{subfigure}

  \caption{
  Extension of Figure~3 (Poisson noise).
  Performance under Poisson noise with varying rate parameter $\lambda$.
}

  \label{fig:app_fig3_poisson}
\end{figure}
\paragraph{Student-$t$ noise.}
We consider additive noise drawn from a Student-$t$ distribution with degrees
of freedom $\nu = 3$,
\begin{equation}
\varepsilon_i \sim t_{\nu}, \quad \nu = 3,
\end{equation}
corresponding to a heavy-tailed noise regime with finite variance.
While the distribution remains substantially heavier-tailed than a Gaussian,
the existence of the second moment places this setting outside the extreme
outlier regime encountered when $\nu \leq 2$.

\begin{figure}[htbp]
  \centering
  \includegraphics[width=0.45\columnwidth]{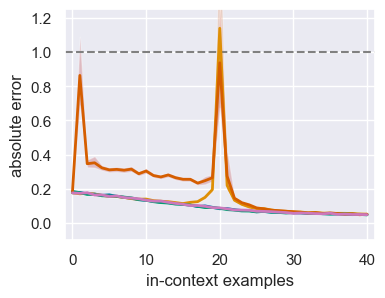}
  \caption{
    Extension of Figure~3 (Student-$t$ noise).
    Performance under heavy-tailed noise beyond the regimes covered by
    standard $\ell_1$ or $\ell_2$ optimality guarantees.
  }
  \label{fig:app_fig3_tstudent}
\end{figure}

Figure~\ref{fig:app_fig3_tstudent} extends the Student-$t$ experiments presented
in the main paper (Figure~3), where $\nu = 2$.
In the $\nu = 2$ case, the noise distribution has infinite variance, and the
resulting extreme outliers induce severe model misspecification for classical
estimators.
In this regime, Transformer-based in-context learning exhibits a clear and
consistent advantage over all baselines, including OLS, Ridge, and
$\ell_1$-based solvers, across a wide range of context lengths.

In contrast, when $\nu = 3$, the noise distribution admits a finite variance,
and the statistical setting becomes substantially closer to the assumptions
under which classical estimators are well behaved.
As shown in Figure~\ref{fig:app_fig3_tstudent}, the Transformer no longer
exhibits a pronounced performance gap, but instead closely matches the
performance of OLS, Ridge, and $\ell_1$-based baselines as the number of
in-context examples increases.
This convergence suggests that once second-order moments are well defined,
classical estimators regain their asymptotic efficiency, and the advantage of
implicit robustness diminishes.

Taken together, these results highlight a sharp transition in behavior across
degrees of freedom.
When the noise variance is infinite ($\nu \leq 2$), Transformer-based
in-context learning demonstrates strong robustness to extreme outliers and
significant gains over fixed-form estimators.
However, once the variance becomes finite ($\nu > 2$), the Transformer's
behavior aligns with that of classical methods, indicating that its advantage
is primarily driven by robustness under severe heavy-tailed noise rather than
by improvements in well-specified, finite-variance regimes.

\end{document}